\documentclass[11pt,a4paper]{article}
\usepackage[hyperref]{emnlp2018}
\usepackage{times}
\usepackage{latexsym}

\usepackage{url}

\usepackage{multirow}
\usepackage{booktabs}

\usepackage{amsmath}
\usepackage{amsfonts}

\DeclareMathOperator{\countop}{count}
\DeclareMathOperator{\score}{score}
\DeclareMathOperator{\nn}{NN}
\DeclareMathOperator{\lex}{lex}

\usepackage{algpseudocode}
\usepackage{algorithm}

\usepackage[utf8]{inputenc}

\usepackage[T1]{fontenc}

\usepackage{graphicx}

\aclfinalcopy 

\title{Unsupervised Statistical Machine Translation}

\author{Mikel Artetxe, Gorka Labaka, Eneko Agirre \\
IXA NLP Group \\
University of the Basque Country (UPV/EHU) \\
\texttt{\{mikel.artetxe,gorka.labaka,e.agirre\}@ehu.eus} \\
}

\date{}

\begin{document}
\maketitle
\begin{abstract}
While modern machine translation has relied on large parallel corpora, a recent line of work has managed to train Neural Machine Translation (NMT) systems from monolingual corpora only \citep{artetxe2018unsupervised,lample2018unsupervised}. Despite the potential of this approach for low-resource settings, existing systems are far behind their supervised counterparts, limiting their practical interest. In this paper, we propose an alternative approach based on phrase-based Statistical Machine Translation (SMT) that significantly closes the gap with supervised systems. Our method profits from the modular architecture of SMT: we first induce a phrase table from monolingual corpora through cross-lingual embedding mappings, combine it with an n-gram language model, and fine-tune hyperparameters through an unsupervised MERT variant. In addition, iterative backtranslation improves results further, yielding, for instance, 14.08 and 26.22 BLEU points in WMT 2014 English-German and English-French, respectively, an improvement of more than 7-10 BLEU points over previous unsupervised systems, and closing the gap with supervised SMT (Moses trained on Europarl) down to 2-5 BLEU points. Our implementation is available at \url{https://github.com/artetxem/monoses}.
\end{abstract}

\section{Introduction}

Neural Machine Translation (NMT) has recently become the dominant paradigm in machine translation \cite{vaswani2017attention}. In contrast to more rigid Statistical Machine Translation (SMT) architectures \cite{koehn2003statistical}, NMT models are trained end-to-end, exploit continuous representations that mitigate the sparsity problem, and overcome the locality problem by making use of unconstrained contexts. Thanks to this additional flexibility, NMT can more effectively exploit large parallel corpora, although SMT is still superior when the training corpus is not big enough \citep{koehn2017six}.

Somewhat paradoxically, while most machine translation research has focused on resource-rich settings where NMT has indeed superseded SMT, a recent line of work has managed to train an NMT system without any supervision, relying on monolingual corpora alone \citep{artetxe2018unsupervised,lample2018unsupervised}. Given the scarcity of parallel corpora for most language pairs, including less-resourced languages but also many combinations of major languages, this research line opens exciting opportunities to bring effective machine translation to many more scenarios. Nevertheless, existing solutions are still far behind their supervised counterparts, greatly limiting their practical usability. For instance, existing unsupervised NMT systems obtain between 15-16 BLEU points in WMT 2014 English-French translation, whereas a state-of-the-art NMT system obtains around 41 \citep{artetxe2018unsupervised,lample2018unsupervised,yang2018unsupervised}.

\begin{figure*}[t] \centering
\includegraphics[width=0.95\textwidth]{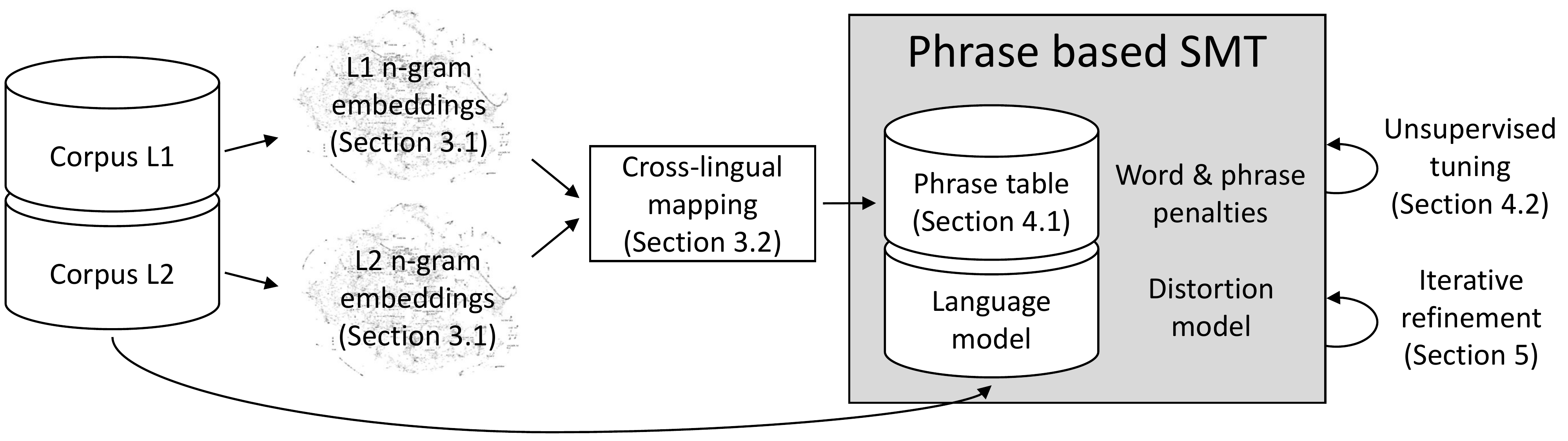}
\caption{Architecture of our system, with references to sections.}
\label{fig:example-init}
\end{figure*}

In this paper, we explore whether the rigid and modular nature of SMT is more suitable for these unsupervised settings, and propose a novel unsupervised SMT system that can be trained on monolingual corpora alone. For that purpose, we present a natural extension of the skip-gram model \cite{mikolov2013distributed} that simultaneously learns word and phrase embeddings, which are then mapped to a cross-lingual space through self-learning \citep{artetxe2018robust}. We use the resulting cross-lingual phrase embeddings to induce a phrase table, and combine it with a language model and a distance-based distortion model to build a standard phrase-based SMT system. The weights of this model are tuned in an unsupervised manner through an iterative Minimum Error Rate Training (MERT) variant, and the entire system is further improved through iterative backtranslation. The architecture of the system is sketched in Figure \ref{fig:example-init}. Our experiments on WMT German-English and French-English datasets show the effectiveness of our proposal, where we obtain improvements above 7-10 BLEU points over previous unsupervised NMT-based approaches, closing the gap with supervised SMT (Moses trained on Europarl) down to 2-5 points.

The remaining of this paper is structured as follows. Section \ref{sec:smt} introduces phrase-based SMT. Section \ref{sec:embeddings} presents our unsupervised approach to learn cross-lingual n-gram embeddings, which are the basis of our proposal. Section \ref{sec:system} describes the proposed unsupervised SMT system itself, while Section \ref{sec:backtranslation} discusses its iterative refinement through backtranslation. Section \ref{sec:experiments} describes the experiments run and the results obtained. Section \ref{sec:related_work} discusses the related work on the topic, and Section \ref{sec:conclusions} concludes the paper.

\section{Background: phrase-based SMT} \label{sec:smt}

While originally motivated as a noisy channel model \citep{brown1990statistical}, phrase-based SMT is now formulated as a log-linear combination of several statistical models that score translation candidates \citep{koehn2003statistical}. The parameters of these scoring functions are estimated independently based on frequency counts, and their weights are then tuned in  a separate validation set. At inference time, a decoder tries to find the translation candidate with the highest score according to the resulting combined model. The specific scoring models found in a standard SMT system are as follows:
\begin{itemize}
\item \textbf{Phrase table}. The phrase table is a collection of source language n-grams and a list of their possible translations in the target language along with different scores for each of them. So as to translate longer sequences, the decoder combines these partial n-gram translations, and ranks the resulting candidates according to their corresponding scores and the rest of scoring functions. In order to build the phrase table, SMT computes word alignments in both directions from a parallel corpus, symmetrizes these alignments using different heuristics \citep{och2003systematic}, extracts the set of consistent phrase pairs, and scores them based on frequency counts. For that purpose, standard SMT uses 4 scores for each phrase table entry: the direct and inverse lexical weightings, which are derived from word level alignments, and the direct and inverse phrase translation probabilities, which are computed at the phrase level.
\item \textbf{Language model}. The language model assigns a probability to a word sequence in the target language. Traditional SMT uses n-gram language models for that, which use simple frequency counts over a large monolingual corpus with back-off and smoothing.
\item \textbf{Reordering model}. The reordering model accounts for different word orders across languages, scoring translation candidates according to the position of each translated phrase in the target language. Standard SMT combines two such models: a distance based distortion model that penalizes deviation from a monotonic order, and a lexical reordering model that incorporates phrase orientation frequencies from a parallel corpus.
\item \textbf{Word and phrase penalties}. The word and phrase penalties assign a fixed score to every generated word and phrase, and are useful to control the length of the output text and the preference for shorter or longer phrases.
\end{itemize}
Having trained all these different models, a tuning process is applied to optimize their weights in the resulting log-linear model, which typically maximizes some evaluation metric in a separate validation parallel corpus. A common choice is to optimize the BLEU score through Minimum Error Rate Training (MERT) \citep{och2003MERT}.

\section{Cross-lingual n-gram embeddings} \label{sec:embeddings}

Section \ref{subsec:phrase_embeddings} presents our proposed extension of skip-gram to learn n-gram embeddings, while Section \ref{subsec:mapping} describes how we map them to a shared space to obtain cross-lingual n-gram embeddings.

\subsection{Learning n-gram embeddings} \label{subsec:phrase_embeddings}
Negative sampling skip-gram takes word-context pairs $(w,c)$, and uses logistic regression to predict whether the pair comes from the true distribution as sampled from the training corpus, or it is one of the $k$ draws from a noise distribution \citep{mikolov2013distributed}:
\[ \log \sigma \left( w \cdot c \right) + \sum_{i=1}^k \mathbb{E}_{c_N \sim P_D} \left[ \log \sigma \left( -w \cdot c_N \right) \right] \]

In its basic formulation, both $w$ and $c$ correspond to words that co-occur within a certain window in the training corpus. So as to learn embeddings for non-compositional phrases like \textit{New York Times} or \textit{Toronto Maple Leafs}, \citet{mikolov2013distributed} propose to merge them into a single token in a pre-processing step. For that purpose, they use a  scoring function based on their co-occurence frequency in the training corpus, with a discounting coefficient $\delta$ that penalizes rare words, and iteratively merge those above a threshold:
\[ \score(w_i, w_j) = \frac{\countop \left( w_i, w_j \right) - \delta}{\countop \left( w_i \right) \times \countop \left( w_j \right)} \]

However, we also need to learn representations for compositional n-grams in our scenario, as there is not always a 1:1 correspondence for n-grams across languages even for compositional phrases.
For instance, the phrase \textit{he will come} would typically be translated as \textit{vendr\'{a}} into Spanish, so one would need to represent the entire phrase as a single unit to properly model this relation. 

One option would be to merge all n-grams regardless of their score, but this is not straightforward given their overlapping nature, which is further accentuated when considering n-grams of different lengths. While we tried to randomly generate multiple consistent segmentations for each sentence and train the embeddings over the resulting corpus, this worked poorly in our preliminary experiments. We attribute this to the complex interactions arising from the stochastic segmentation (e.g. the co-occurrence distribution changes radically, even for unigrams), severely accentuating the sparsity problem, among other issues.

As an alternative approach, we propose a generalization of skip-gram that learns n-gram embeddings on-the-fly, and has the desirable property of unigram invariance: our proposed model learns the exact same embeddings as the original skip-gram for unigrams, while simultaneously learning additional embeddings for longer n-grams. This way, for each word-context pair $(w,c)$ at distance $d$ within the given window, we update their corresponding embeddings $w$ and $c$ with the usual negative sampling loss. In addition to that, we look at all n-grams $p$ of different lengths that are at the same distance $d$, and for each pair $(p,c)$, we update the embedding $p$ through negative sampling. In order to enforce unigram invariance, the context $c$ and negative samples $c_N$, which always correspond to unigrams, are not updated for $(p,c)$. This allows to naturally learn n-gram embeddings according to their co-occurrence patterns as modeled by skip-gram, without introducing subtle interactions that affect its fundamental behavior.

We implemented the above procedure as an extension of \textit{word2vec}, and use it to train monolingual n-gram embeddings with a window size of 5, 300 dimensions, 10 negative samples, 5 iterations and subsampling disabled. So as to keep the model size within a reasonable limit, we restrict the vocabulary to the most frequent 200,000 unigrams, 400,000 bigrams and 400,000 trigrams.

\subsection{Cross-lingual mapping} \label{subsec:mapping}

Cross-lingual mapping methods take independently trained word embeddings in two languages, and learn a linear transformation to map them to a shared cross-lingual space \citep{mikolov2013exploiting,artetxe2018generalizing}. Most mapping methods are supervised, and rely on a bilingual dictionary, typically in the range of a few thousand entries, although a recent line of work has managed to achieve comparable results in a fully unsupervised manner based on either self-learning \citep{artetxe2017learning,artetxe2018robust} or adversarial training \citep{zhang2017adversarial,zhang2017earth,conneau2018word}.

In our case, we use the method of \citet{artetxe2018robust} to map the n-gram embeddings to a shared cross-lingual space using their open source implementation VecMap\footnote{\url{https://github.com/artetxem/vecmap}}. Originally designed for word embeddings, this method builds an initial mapping by connecting the intra-lingual similarity distribution of embeddings in different languages, and iteratively improves this solution through self-learning. The method applies a frequency-based vocabulary cut-off, learning the mapping over the 20,000 most frequent words in each language. We kept this cut-off to learn the mapping over the most frequent 20,000 unigrams, and then apply the resulting mapping to the entire embedding space, including longer n-grams.

\section{Unsupervised SMT} \label{sec:system}

As discussed in Section \ref{sec:smt}, phrase-based SMT follows a modular architecture that combines several scoring functions through a log-linear model. Among the scoring functions found in standard SMT systems, the distortion model and word/phrase penalties are parameterless, while the language model is trained on monolingual corpora, so they can all be directly integrated into our unsupervised system. From the remaining models, typically trained on parallel corpora, we decide to leave the lexical reordering out, as the distortion model already accounts for word reordering. As for the phrase table, we learn cross-lingual n-gram embeddings as discussed in Section \ref{sec:embeddings}, and use them to induce and score phrase translation pairs as described next (Section \ref{subsec:phrase_table_induction}). Finally, we tune the weights of the resulting log-linear model using an unsupervised procedure based on backtranslation (Section \ref{subsec:tuning}).

Unless otherwise specified, we use Moses\footnote{\url{http://www.statmt.org/moses/}} with default hyperparameters to implement these different components of our system. We use KenML \citep{heafield2013scalable}, bundled in Moses by default, to estimate our 5-gram language model with modified Kneser-Ney smoothing, pruning n-grams longer than 3 with a single occurrence.

\subsection{Phrase table induction} \label{subsec:phrase_table_induction}

Given the lack of a parallel corpus from which to \textbf{extract phrase translation pairs}, every n-gram in the target language could be taken as a potential translation candidate for each n-gram in the source language. So as to keep the size of the phrase table within a reasonable limit, we train cross-lingual phrase embeddings as described in Section \ref{sec:embeddings}, and limit the translation candidates for each source phrase to its 100 nearest neighbors in the target language.

In order to estimate their corresponding \textbf{phrase translation probabilities}, we apply the softmax function over the cosine similarities of their respective embeddings. More concretely, given the source language phrase $\bar{e}$ and the translation candidate $\bar{f}$, their direct phrase translation probability is computed as follows\footnote{The inverse phrase translation probability $\phi ( \bar{e} | \bar{f} )$ is defined analogously.}:
\[ \phi ( \bar{f} | \bar{e} ) = \frac{ \cos( \bar{e}, \bar{f} ) / \tau}{\sum_{\bar{f'}} \cos ( \bar{e}, \bar{f'} ) / \tau}\]
Note that, in the above formula, $\bar{f'}$ iterates across all target language embeddings, and $\tau$ is a constant temperature parameter that controls the confidence of the predictions. In order to tune it, we induce a dictionary over the cross-lingual embeddings themselves with nearest neighbor retrieval, and use maximum likelihood estimation over it. However, inducing the dictionary in the same direction as the probability predictions leads to a degenerated solution (softmax approximates the hard maximum underlying nearest neighbor as $\tau$ approaches 0), so we induce the dictionary in the opposite direction and apply maximum likelihood estimation over it:
\[ \min_\tau \sum_{\bar{f}} \log \phi ( \bar{f} | \nn_{\bar{e}} (\bar{f}) ) + \sum_{\bar{e}} \log \phi ( \bar{e} | \nn_{\bar{f}} (\bar{e}) ) \]
So as to optimize $\tau$, we use Adam with a learning rate of 0.0003 and a batch size of 200, implemented in PyTorch.

In order to compute the \textbf{lexical weightings}, we align each word in the target phrase with the one in the source phrase most likely generating it, and take the product of their respective translation probabilities:
\[ \lex ( \bar{f} | \bar{e} ) = \prod_i \max \left( \epsilon, \max_j w (\bar{f}_i | \bar{e}_j) \right) \]
The constant $\epsilon$ guarantees that each target language word will get a minimum probability mass, which is useful to model NULL alignments. In our experiments, we set $\epsilon=0.001$, which we find to work well in practice. Finally, the word translation probabilities $w (\bar{f}_i | \bar{e}_j)$ are computed using the same formula defined for phrase translation probabilities (see above), with the difference that the partition function goes over unigrams only.

\subsection{Unsupervised tuning} \label{subsec:tuning}

As discussed in Section \ref{sec:smt}, standard SMT uses MERT over a small parallel corpus to tune the weights of the different scoring functions combined through its log-linear model. Given that we only have access to monolingual corpora in our scenario, we propose to generate a synthetic parallel corpus through backtranslation \citep{sennrich2016improving} and apply MERT tuning over it, iteratively repeating the process in both directions (see Algorithm \ref{alg:tuning}). For that purpose, we reserve a random subset of 10,000 sentences from each monolingual corpora, and run the proposed algorithm over them for 10 iterations, which we find to be enough for convergence.

\begin{algorithm}[t]
\begin{algorithmic}[1]
\item[\textbf{Input:} $m_{s \rightarrow t}$ (source-to-target models)]
\item[\textbf{Input:} $m_{t \rightarrow s}$ (target-to-source models)]
\item[\textbf{Input:} $c_{s}$ (source validation corpus)]
\item[\textbf{Input:} $c_{t}$ (target validation corpus)]
\item[\textbf{Output:} $w_{s \rightarrow t}$ (source-to-target weights)]
\item[\textbf{Output:} $w_{t \rightarrow s}$ (target-to-source weights)]
\State $w_{t \rightarrow s} \gets$ \Call{default\_weights}{}
\Repeat
  \State $bt_s \gets$ \Call{translate}{$m_{t \rightarrow s}$, $w_{t \rightarrow s}$, $c_t$}
  \State $w_{s \rightarrow t} \gets$ \Call{mert}{$m_{s \rightarrow t}$, $bt_s$, $c_t$}
  \State $bt_t \gets$ \Call{translate}{$m_{s \rightarrow t}$, $w_{s \rightarrow t}$, $c_s$}
  \State $w_{t \rightarrow s} \gets$ \Call{mert}{$m_{t \rightarrow s}$, $bt_t$, $c_s$}
\Until{convergence}
\end{algorithmic}
\caption{Unsupervised tuning}
\label{alg:tuning}
\end{algorithm}

\section{Iterative refinement} \label{sec:backtranslation}
The procedure described in Section \ref{sec:system} suffices to train an SMT system from monolingual corpora which, as shown by our experiments in Section \ref{sec:experiments}, already outperforms previous unsupervised systems. Nevertheless, our proposed method still makes important simplifications that could compromise its potential performance: it does not use any lexical reordering model, its phrase table is limited by the underlying embedding vocabulary (e.g. it does not include phrases longer than trigrams, see Section \ref{subsec:phrase_embeddings}), and the phrase translation probabilities and lexical weightings are estimated based on cross-lingual embeddings.

In order to overcome these limitations, we propose an iterative refinement procedure based on backtranslation \citep{sennrich2016improving}. More concretely, we generate a synthetic parallel corpus by translating the monolingual corpus in one of the languages with the initial system, and train and tune a standard SMT system over it in the opposite direction. Note that this new system does not have any of the initial restrictions: the phrase table is built and scored using standard word alignment with an unconstrained vocabulary, and a lexical reordering model is also learned. Having done that, we use the resulting system to translate the monolingual corpus in the other language, and train another SMT system over it in the other direction. As detailed in Algorithm \ref{alg:refinement}, this process is repeated iteratively until some convergence criterion is met.

\begin{algorithm}[t]
\begin{algorithmic}[1]
\item[\textbf{Input:} $c_{s}$ (source language corpus)]
\item[\textbf{Input:} $c_{t}$ (target language corpus)]
\item[\textbf{Input/Output:} $m_{t \rightarrow s}$ (target-to-source models)]
\item[\textbf{Input/Output:} $w_{t \rightarrow s}$ (target-to-source weights)]
\item[\textbf{Output:} $m_{s \rightarrow t}$ (source-to-target models)]
\item[\textbf{Output:} $w_{s \rightarrow t}$ (source-to-target weights)]
\State $train_{s}, val_{s} \gets$ \Call{split}{$c_s$}
\State $train_{t}, val_{t} \gets$ \Call{split}{$c_t$}
\Repeat
  \State $btt_s \gets$ \Call{translate}{$m_{t \rightarrow s}$, $w_{t \rightarrow s}$, $train_t$}
  \State $btv_s \gets$ \Call{translate}{$m_{t \rightarrow s}$, $w_{t \rightarrow s}$, $val_t$}
  \State $m_{s \rightarrow t} \gets$ \Call{train}{$btt_s$, $train_t$}
  \State $w_{s \rightarrow t} \gets$ \Call{mert}{$m_{s \rightarrow t}$, $btv_s$, $val_t$}
  \State $btt_t \gets$ \Call{translate}{$m_{s \rightarrow t}$, $w_{s \rightarrow t}$, $train_s$}
  \State $btv_t \gets$ \Call{translate}{$m_{s \rightarrow t}$, $w_{s \rightarrow t}$, $val_s$}
  \State $m_{t \rightarrow s} \gets$ \Call{train}{$btt_t$, $train_s$}
  \State $w_{t \rightarrow s} \gets$ \Call{mert}{$m_{t \rightarrow s}$, $btv_t$, $val_s$}
\Until{convergence}
\end{algorithmic}
\caption{Iterative refinement}
\label{alg:refinement}
\end{algorithm}

\begin{table*}[t]
\begin{center}
  \begin{tabular}{lccccccc}
    \toprule
    & \multicolumn{4}{c}{\bf WMT-14} & & \multicolumn{2}{c}{\bf WMT-16} \\
    \cmidrule{2-5} \cmidrule{7-8}
    & \multicolumn{1}{c}{\bf FR-EN} & \multicolumn{1}{c}{\bf EN-FR} & \multicolumn{1}{c}{\bf DE-EN} & \multicolumn{1}{c}{\bf EN-DE} & & \multicolumn{1}{c}{\bf DE-EN} & \multicolumn{1}{c}{\bf EN-DE} \\
    \midrule
    \citet{artetxe2018unsupervised} & 15.56 & 15.13 & 10.21 & 6.55 & & - & - \\
    \citet{lample2018unsupervised} & 14.31 & 15.05 & - & - & & 13.33 & 9.64 \\
    \citet{yang2018unsupervised} & 15.58 & 16.97 & - & - & & 14.62 & 10.86 \\
    \midrule
    Proposed system & \bf 25.87 & \bf 26.22 & \bf 17.43 & \bf 14.08 & & \bf 23.05 & \bf 18.23 \\
    \bottomrule
  \end{tabular}
\end{center}
\caption{Results of the proposed method in comparison to existing unsupervised NMT systems (BLEU).}
\label{tab:results_sota}
\end{table*}

While this procedure would be expected to produce a more accurate model at each iteration, it also happens to be very expensive computationally. In order to accelerate our experiments, we use a random subset of 2 million sentences from each monolingual corpus for training\footnote{Note that we reuse the original language model, which is trained in the full corpus.}, in addition to the 10,000 separate sentences that are held out as a validation set for MERT tuning, and perform a fixed number of 3 iterations of the above algorithm. Moreover, we use FastAlign \citep{dyer2013simple} instead of GIZA++ to make word alignment faster. Other than that, training over the synthetic parallel corpus is done through standard Moses tools with default settings.

\section{Experiments and results} \label{sec:experiments}

In order to make our experiments comparable to previous work, we use the French-English and German-English datasets from the WMT 2014 shared task. As discussed throughout the paper, our system is trained on monolingual corpora alone, so we take the concatenation of all News Crawl monolingual corpora from 2007 to 2013 as our training data, which we tokenize and truecase using standard Moses tools. The resulting corpus has 749 million tokens in French, 1,606 million tokens in German, and 2,109 million tokens in English. Following common practice, the systems are evaluated in newstest2014 using tokenized BLEU scores as computed by the \texttt{multi-bleu.perl} script included in Moses. In addition to that, we also report results in German-English newstest2016 (from WMT 2016), as this was used by some previous work in unsupervised NMT \citep{lample2018unsupervised,yang2018unsupervised}\footnote{Note that we use the same model trained in WMT 2014 for these experiments, so it is likely that our results could be further improved by using the more extensive data from WMT 2016.}. So as to be faithful to our target scenario, we did not use any parallel data in these language pairs, not even for development purposes. Instead, we ran all our preliminary experiments on WMT Spanish-English data, where we made all development decisions.

We present the results of our final system in comparison to other previous work in Section \ref{subsec:results_main}. Section \ref{subsec:results_ablation} then presents an ablation study of our proposed method, where we analyze the contribution of its different components. Section \ref{subsec:results_supervised} compares the obtained results to those of different supervised systems, analyzing the effect of some of the inherent limitations of our method in a standard phrase-based SMT system. Finally, Section \ref{subsec:results_examples} presents some translation examples from our system.

\subsection{Main results} \label{subsec:results_main}

We report the results obtained by our proposed system in Table \ref{tab:results_sota}. As it can be seen, our system obtains the best published results by a large margin, surpassing previous unsupervised NMT systems by around 10 BLEU points in French-English (both directions), and more than 7 BLEU points in German-English (both directions and datasets).

This way, while previous progress in the task has been rather incremental \citep{yang2018unsupervised}, our work represents an important step towards high-quality unsupervised machine translation, with improvements over 50\% in all cases. This suggests that, in contrast to previous NMT-based approaches, phrase-based SMT may provide a more suitable framework for unsupervised machine translation, which is in line with previous results in low-resource settings \citep{koehn2017six}.

\begin{table*}[t]
\begin{center}
  \begin{tabular}{lccccccc}
    \toprule
    & \multicolumn{4}{c}{\bf WMT-14} & & \multicolumn{2}{c}{\bf WMT-16} \\
    \cmidrule{2-5} \cmidrule{7-8}
    & \multicolumn{1}{c}{\bf FR-EN} & \multicolumn{1}{c}{\bf EN-FR} & \multicolumn{1}{c}{\bf DE-EN} & \multicolumn{1}{c}{\bf EN-DE} & & \multicolumn{1}{c}{\bf DE-EN} & \multicolumn{1}{c}{\bf EN-DE} \\
    \midrule
    Unsupervised SMT & 21.16 & 20.13 & 13.86 & 10.59 & & 18.01 & 13.22 \\
    + unsupervised tuning & 22.17 & 22.22 & 14.73 & 10.64 & & 18.21 & 13.12 \\
    + iterative refinement (it1) & 24.81 & 26.53 & 16.01 & 13.45 & & 20.76 & 16.94 \\
    + iterative refinement (it2) & \bf 26.13 & \bf 26.57 & 17.30 & 13.95 & & 22.80 & 18.18 \\
    + iterative refinement (it3) & 25.87 & 26.22 & \bf 17.43 & \bf 14.08 & & \bf 23.05 & \bf 18.23 \\
    \bottomrule
  \end{tabular}
\end{center}
\caption{Ablation results (BLEU). The last row corresponds to our full system. Refer to the text for more details.}
\label{tab:results_ablation}
\end{table*}

\begin{table*}[t]
\begin{center}
  \begin{tabular}{clccccccc}
    \toprule
    & & \multicolumn{4}{c}{\bf WMT-14} & & \multicolumn{2}{c}{\bf WMT-16} \\
    \cmidrule{3-6} \cmidrule{8-9}
    & & \multicolumn{1}{c}{\bf FR-EN} & \multicolumn{1}{c}{\bf EN-FR} & \multicolumn{1}{c}{\bf DE-EN} & \multicolumn{1}{c}{\bf EN-DE} & & \multicolumn{1}{c}{\bf DE-EN} & \multicolumn{1}{c}{\bf EN-DE} \\
    \midrule
    \bf \multirow{7}{*}{Supervised} & NMT (transformer) & - & 41.8 & - & 28.4 & & - & - \\
    \cmidrule{2-9}
    & WMT best & 35.0 & 35.8 & 29.0 & 20.6 & & 40.2 & 34.2 \\
    \cmidrule{2-9}
    & SMT (europarl) & 30.61 & 30.82 & 20.83 & 16.60 & & 26.38 & 22.12 \\
    & + w/o lexical reord. & 30.54 & 30.33 & 20.37 & 16.34 & & 25.99 & 22.20 \\
    & + constrained vocab. & 30.04 & 30.10 & 19.91 & 16.32 & & 25.66 & 21.53 \\
    & + unsup. tuning & 29.32 & 29.46 & 17.75 & 15.45 & & 23.35 & 19.86 \\
    \midrule
    \bf Unsup. & Proposed system & 25.87 & 26.22 & 17.43 & 14.08 & & 23.05 & 18.23 \\
    \bottomrule
  \end{tabular}
\end{center}
\caption{Results of the proposed method in comparison to supervised systems (BLEU). Transformer results reported by \citet{vaswani2017attention}. SMT variants are incremental (e.g. 2nd includes 1st). Refer to the text for more details.}
\label{tab:results_supervised}
\end{table*}

\subsection{Ablation analysis} \label{subsec:results_ablation}

We present ablation results of our proposed system in Table \ref{tab:results_ablation}. The first row corresponds to the initial system with our induced phrase table (Section \ref{subsec:phrase_table_induction}) and default weights as used by Moses, whereas the second row uses our unsupervised MERT procedure to tune these weights (Section \ref{subsec:tuning}). The remaining rows represent different iterations of our refinement procedure (Section \ref{sec:backtranslation}), which uses backtranslation to iteratively train a standard SMT system from a synthetic parallel corpus.

The results show that the initial system with default weights (first row) is already better than previous unsupervised NMT systems (Table \ref{tab:results_sota}) by a substantial margin (2-6 BLEU points). Our unsupervised tuning procedure further improves results, bringing an improvement of over 1 BLEU point in both French-English directions, although its contribution is somewhat weaker for German-to-English (almost 1 BLEU point in WMT 2014 but only 0.2 in WMT 2016), and does not make any difference for English-to-German.

The proposed iterative refinement method has a much stronger positive effect, with improvements over 2.5 BLEU points in all cases, and up to 5 BLEU points in some. Most gains come in the first iteration, while the second iteration brings weaker improvements and the algorithm seems to converge in the third iteration, with marginal improvements for German-English and a small drop in performance for French-English.

\begin{table*}[t]
\begin{small}
\begin{center}
  \begin{tabular}{p{4.8cm}p{4.8cm}p{4.8cm}}
    \toprule
    \bf Source & \bf Reference & \bf Proposed system \\
    \midrule
    D'autres révélations ont fait état de documents divulgués par Snowden selon lesquels la NSA avait intercepté des données et des communications émanant du téléphone portable de la chancelière allemande Angela Merkel et de ceux de 34 autres chefs d'État. & Other revelations cited documents leaked by Snowden that the NSA monitored German Chancellor Angela Merkel's cellphone and those of up to 34 other world leaders. & Other disclosures have reported documents disclosed by Snowden suggested the NSA had intercepted communications and data from the mobile phone of German Chancellor Angela Merkel and those of 32 other heads of state. \\
    \midrule
	La NHTSA n'a pas pu examiner la lettre d'information aux propriétaires en raison de l'arrêt de 16 jours des activités gouvernementales, ce qui a ralenti la croissance des ventes de véhicules en octobre. & NHTSA could not review the owner notification letter due to the 16-day government shutdown, which tempered auto sales growth in October. & The NHTSA could not consider the letter of information to owners because of halting 16-day government activities, which slowed the growth in vehicle sales in October. \\
    \midrule
    Le M23 est né d'une mutinerie, en avril 2012, d'anciens rebelles, essentiellement tutsi, intégrés dans l'armée en 2009 après un accord de paix. & The M23 was born of an April 2012 mutiny by former rebels, principally Tutsis who were integrated into the army in 2009 following a peace agreement. & M23 began as a mutiny in April 2012, former rebels, mainly Tutsi integrated into the national army in 2009 after a peace deal. \\
    \midrule
	Tunks a déclaré au Sunday Telegraph de Sydney que toute la famille était <<extrêmement préoccupée>> du bien-être de sa fille et voulait qu'elle rentre en Australie. & Tunks told Sydney's Sunday Telegraph the whole family was ``extremely concerned'' about his daughter's welfare and wanted her back in Australia. & Tunks told The Times of London from Sydney that the whole family was ``extremely concerned'' of the welfare of her daughter and wanted it to go in Australia. \\
    \bottomrule
  \end{tabular}
\end{center}
\end{small}
\caption{Randomly chosen translation examples from French$\rightarrow$English newstest2014.} \label{tab:examples}
\end{table*}

\subsection{Comparison with supervised systems} \label{subsec:results_supervised}

So as to put our results into perspective, Table \ref{tab:results_supervised} comprises the results of different supervised methods in the same test sets. More concretely, we report the results of the Transformer \cite{vaswani2017attention}, an NMT system based on self-attention that is the current state-of-the-art in machine translation, along with the scores obtained by the best performing system in each WMT shared task at the time, and those of a standard phrase-based SMT system trained on Europarl and tuned on newstest2013 using Moses. We also report the effect of removing lexical reordering from the latter as we do in our initial system (Section \ref{sec:system}), restricting the vocabulary to the most frequent unigram, bigram and trigrams as we do when training our embeddings (Section \ref{sec:embeddings}), and using our unsupervised tuning procedure over a subset of the monolingual corpus (Section \ref{subsec:tuning}) instead of using standard MERT tuning over newstest2013.

Quite surprisingly, our proposed system, trained exclusively on monolingual corpora, is relatively close to a comparable phrase-based SMT system trained on Europarl, with differences below 5 BLEU points in all cases and as little as 2.5 in some. Note that both systems use the exact same language model trained on News Crawl, making them fully comparable in terms of the monolingual corpora they have access to. While more of a baseline than the state-of-the-art, note that Moses+Europarl is widely used as a reference system in machine translation. As such, we think that our results are very encouraging, as they show that our fully unsupervised system is already quite close to this competitive baseline.

In addition to that, the results for the constrained variants of this SMT system justify some of the simplifications required by our approach. In particular, removing lexical reordering and constraining the phrase table to the most frequent n-grams, as we do for our initial system, has a relatively small effect, with a drop of less than 1 BLEU point in all cases, and as little as 0.28 in some. Replacing standard MERT tuning with our unsupervised variant does cause a considerable drop in performance, although it is below 2.5 BLEU points even in the worst case, and our unsupervised tuning method is still better than using default weights as reported in Table \ref{tab:results_ablation}. This shows the importance of tuning in SMT, suggesting that these results could be further improved if one had access to a small parallel corpus for tuning.

\subsection{Qualitative results} \label{subsec:results_examples}

Table \ref{tab:examples} shows some of the translations produced by the proposed system for French$
\rightarrow$English. Note that these examples where randomly taken from the test set, so they should be representative of the general behavior of our approach.

While the examples reveal certain adequacy issues (e.g. \textit{The Times of London from Sidney} instead of \textit{Sydney's Sunday Telegraph}), and the produced output is not perfectly grammatical (e.g. \textit{go in Australia}), our translations are overall quite accurate and fluent, and one could get a reasonable understanding of the original text from them. This suggests that unsupervised machine translation can indeed be a usable alternative in low resource settings.

\section{Related work} \label{sec:related_work}

Similar to our approach, statistical decipherment also attempts to build machine translation systems from monolingual corpora. For that purpose, existing methods treat the source language as ciphertext, and model its generation through a noisy channel model involving two steps: the generation of the original English sentence and the probabilistic replacement of the words in it \citep{ravi2011deciphering,dou2012large}. The English generative process is modeled using an n-gram language model, and the channel model parameters are estimated using either expectation maximization or Bayesian inference. Subsequent work has attempted to enrich these models with additional information like syntactic knowledge \citep{dou2013dependency} and word embeddings \citep{dou2015unifying}. Nevertheless, these systems work in a word-by-word basis and have only been shown to work in limited settings, being often evaluated in word-level translation. In contrast, our method builds a fully featured phrase-based SMT system, and achieves competitive performance in a standard machine translation task.

More recently, \citet{artetxe2018unsupervised} and \citet{lample2018unsupervised} have managed to train a standard attentional encoder-decoder NMT system from monolingual corpora alone. For that purpose, they use a shared encoder for both languages with pre-trained cross-lingual embeddings, and train the entire system using a combination of denoising, backtranslation and, in the case of \citet{lample2018unsupervised}, adversarial training. This method was further improved by \citet{yang2018unsupervised},
who use a separate encoder for each language, sharing only a subset of their parameters, and incorporate two generative adversarial networks. However, our results in Section \ref{subsec:results_main} show that our SMT-based approach obtains substantially better results.

Our method is also connected to some previous approaches to improve machine translation using monolingual corpora. In particular, the generation of a synthetic parallel corpus through backtranslation \citep{sennrich2016improving}, which is a key component of our unsupervised tuning and iterative refinement procedures, has been previously used to improve NMT. In addition, there have been several proposals to extend the phrase table of SMT systems by inducing translation candidates and/or scores from monolingual corpora, using either statistical decipherment methods \citep{dou2012large,dou2013dependency} or cross-lingual embeddings \citep{zhao2015learning,wang2016bilingual}. While all these methods exploit monolingual corpora to enhance an existing machine translation system previously trained on parallel corpora, our approach learns a fully featured phrase-based SMT system from monolingual corpora alone.

\section{Conclusions and future work} \label{sec:conclusions}

In this paper, we propose a novel unsupervised SMT system that can be trained on monolingual corpora alone. For that purpose, we extend the skip-gram model \citep{mikolov2013distributed} to simultaneously learn word and phrase embeddings, and map them to a cross-lingual space adapting previous unsupervised techniques \citep{artetxe2018robust}. The resulting cross-lingual phrase embeddings are used to induce a phrase table, which coupled with an n-gram language model and distance-based distortion yields an unsupervised phrase-based SMT system. We further improve results tuning the weights with our unsupervised MERT variant, and obtain additional improvements re-training the entire system through iterative backtranslation. Our implementation is available as an open source project at \url{https://github.com/artetxem/monoses}.

Our experiments on standard WMT French-English and German-English datasets confirm the effectiveness of our proposal, where we obtain improvements above 10 and 7 BLEU points over previous NMT-based approaches, respectively, closing the gap with supervised SMT (Moses trained on Europarl) down to 2-5 points.

In the future, we would like to extend our approach to semi-supervised scenarios with small parallel corpora, which we expect to be particularly helpful for tuning purposes. Moreover, we would like to try a hybrid approach with NMT, using our unsupervised SMT system to generate a synthetic parallel corpus and training an NMT system over it through iterative backtranslation.

\section*{Acknowledgments}

This research was partially supported by the Spanish MINECO (TUNER TIN2015-65308-C5-1-R, MUSTER PCIN-2015-226 and TADEEP TIN2015-70214-P, cofunded by EU FEDER), the UPV/EHU (excellence research group), and the NVIDIA GPU grant program. Mikel Artetxe enjoys a doctoral grant from the Spanish MECD.

\bibliography{emnlp2018}
\bibliographystyle{acl_natbib_nourl}

\end{document}